\documentclass{article}

\usepackage{template2/aaai23}  

\usepackage[utf8]{inputenc} 
\usepackage[T1]{fontenc}    
\usepackage{hyperref}       
\usepackage{url}            
\usepackage{booktabs}       
\usepackage{amsfonts}       
\usepackage{nicefrac}       
\usepackage{microtype}      
\usepackage{xcolor}         

\usepackage{subfiles} 
\usepackage{graphicx}
\graphicspath{{figures/}{../figures/}}
\usepackage{subfig}

\usepackage{float}
\restylefloat{figure}

\usepackage{bbding}

\title{Eloss: An Interpretability Amplifier of 3D Object Detection Network for Intelligent Driving}

\author{%
    Haobo Yang, Shiyan Zhang, Zhuoyi Yang, Xinyu Zhang\thanks{Corresponding author} \\
    \textbf{Jilong Guo, Zongyou Yang, Jun Li} \\
    the State Key Laboratory of Automotive Safety and Energy, \\
    and the School of Vehicle and Mobility\\
    Tsinghua University\\
    \texttt{s1911593@ed.ac.uk, zshiyan@bupt.edu.cn,} \\
    \texttt{zhuoyiyang03241811@tju.edu.cn, xyzhang@tsinghua.edu.cn,} \\
    \texttt{2018040160@buct.edu.cn, ucabz77@ucl.ac.uk} \\
    \texttt{lijun19580326@126.com}
}

\begin{document}
\maketitle

\begin{abstract}

    With the increasing complexity of the traffic environment, the significance of safety perception in intelligent driving is intensifying. Traditional methods in the field of intelligent driving perception rely on deep learning, which suffers from limited interpretability, often described as a "black box." This paper introduces a novel type of loss function, termed "Entropy Loss," along with an innovative training strategy. Entropy Loss is formulated based on the functionality of feature compression networks within the perception model. Drawing inspiration from communication systems, the information transmission process in a feature compression network is expected to demonstrate steady changes in information volume and a continuous decrease in information entropy. By modeling network layer outputs as continuous random variables, we construct a probabilistic model that quantifies changes in information volume. Entropy Loss is then derived based on these expectations, guiding the update of network parameters to enhance network interpretability. Our experiments indicate that the Entropy Loss training strategy accelerates the training process. Utilizing the same 60 training epochs, the accuracy of 3D object detection models using Entropy Loss on the KITTI test set improved by up to 4.47\% compared to models without Entropy Loss, underscoring the method's efficacy. The implementation code is available at \url{https://github.com/yhbcode000/Eloss-Interpretability}.

\end{abstract}

\section{Introduction}

    As urban transportation evolves, intelligent driving emerges as an inevitable trend and a cornerstone of future mobility \cite{freudendal2019smart}. Among its core capabilities, 3D object detection is pivotal \cite{zhi2017lightnet}. Traditional detection methods relying solely on a single sensor often fall short in providing sufficient data for accurate object detection \cite{xu2018multi}. In response, the integration of multisensor data through collaborative detection methods has become crucial. The exponential growth of deep learning has enabled effective fusion of multimodal sensory data, significantly enhancing the precision of 3D object detection \cite{wu2021multi}. This technological progression has transformed the perception systems from single-sensor to sophisticated multisensor collaborative frameworks \cite{chen2019cooper, song2021collaborative}.

    \begin{figure}
        \centering
        \includegraphics[width=\columnwidth]{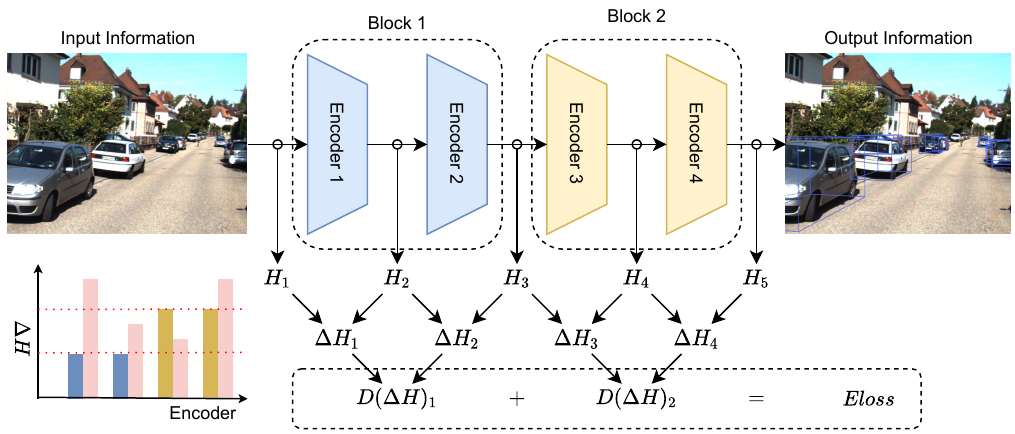}
        \caption{An detailed Illustration of our proposed Eloss method.}
        \label{fig eloss.}
    \end{figure}

    The ongoing research in multimodal perception for intelligent driving predominantly focuses on the architectural design of neural networks \cite{feng2020deep}. The widespread adoption of intelligent vehicles, however, hinges on their transparency, credibility, and regulatory compliance \cite{lim2018autonomous}. It underscores the necessity of developing interpretable multimodal networks, not just for operational efficiency but also to foster public trust and facilitate broader acceptance.

    The opacity of deep learning models, often referred to as the "black box," poses significant challenges. Addressing these challenges is critical not only for enhancing the understanding of neural network operations but also for ensuring the safety and reliability of autonomous vehicles. Researchers are tasked with overcoming several hurdles:

    \begin{itemize}
        \item Crafting network architectures that are optimally suited for perception tasks while providing insights into their operational mechanisms.
        \item Evaluating the effectiveness and logical coherence of features extracted by deep learning models remains a complex issue \cite{samek2016evaluating}.
        \item Given the complexity of scenarios encountered in large-scale intelligent driving, leveraging network interpretability effectively poses substantial challenges \cite{nguyen2020autonomous}.
    \end{itemize}

    This paper approaches the design of perception systems in intelligent driving from the vantage point of communication models \cite{zou2022novel}, establishing a theoretical framework that elucidates the fundamental mechanisms of neural networks in this context. Our contributions are manifold:

    First, we draw upon the principles of source encoding from communication theory to construct a model where the information flow within each layer of a feature compression network is consistent and predictable \cite{jones_2000_information}.

    Second, we introduce a probabilistic model utilizing a continuous random variable $X$, which enables the quantification of information entropy changes within the network, thus providing a clearer understanding of data transformation processes.

    Third, we propose the Entropy Loss function, an innovative approach that not only facilitates the real-time adjustment of network parameters but also integrates seamlessly with existing neural network training methodologies, enhancing both interpretability and training efficacy.

    By addressing these challenges, our work lays a solid foundation for the development of highly reliable and transparent perception systems essential for the safe operation of intelligent vehicles.

\section{Related Works}

    \subsection{Interpretability of Neural Networks}
    
        Interpretability in neural networks is increasingly recognized as a crucial aspect for applications in areas demanding high reliability and transparency, such as intelligent driving. Interpretation methods typically focus on three critical dimensions: model performance, computational cost, and interpretability itself \cite{carvalho2019machine}.
    
        Simonyan et al.\cite{simonyan2013deep} introduced two seminal techniques aimed at enhancing model transparency: “class model visualization” to identify the most representative image for a given class after model training, and “class saliency visualization” which helps in interpreting the contributions of different image regions to the final decision made by the network. While these methods have advanced real-time interpretative frameworks, they necessitate additional computational layers, significantly increasing the training burden \cite{do2018real}. Further extending the realm of interpretability, Li et al. \cite{li2021rts3d} developed a multimodal model that not only automates the diagnosis of mental illnesses but also provides explanations for its diagnostic decisions.
    
        Despite these advances, WuFei et al. \cite{wu_2019_interpretability} argue that the field is still nascent, with statistical analyses beginning to uncover how deep learning models function internally. These methods, effective for the interpretation of individual images or simpler tasks, often struggle under the computational load required by the larger, more complex networks needed for tasks like intelligent driving.
    
        The interpretability of deep neural networks, particularly those based on fusion models, remains limited. Uncontrolled fusion directions can lead to unpredictable, and sometimes suboptimal, performance compared to single-modality systems. This unpredictability underscores the necessity for robust, interpretable multimodal networks that can reliably integrate and interpret diverse data streams, thus enhancing the decision-making process in intelligent driving systems.
    
        Current research illustrates the pivotal role of deep learning interpretability, particularly how it can either obstruct or facilitate the practical deployment of AI technologies. However, studies focused on the mechanisms of multimodal fusion are still exploratory, lacking comprehensive demonstrations and rigorous experimental validations. As intelligent driving systems evolve, the development of interpretable models that can effectively combine and rationalize data from various sensors will become critical. This necessitates a deeper theoretical understanding, coupled with advanced methodologies that can mitigate the inherent complexity of these systems.
    
        Future research directions might include the development of low-complexity interpretative frameworks that do not compromise computational efficiency. Additionally, the exploration of novel neural architectures that inherently facilitate interpretability, such as transparent neural networks or inherently interpretable models, could pave the way for safer and more reliable intelligent driving technologies. Such advancements could dramatically enhance the deployment potential of autonomous vehicles, aligning with societal expectations and regulatory standards.

    \subsection{Shannon's Source Coding in the Communication Model}
    
        In the realm of communications, networks are inherently interpretable, structured on the robust theoretical foundations provided by information theory. This allows for the application of quantitative metrics to assess network reliability. Recent advances have seen deep neural networks employ joint source-channel coding \cite{kurka_2020_deepjsccf,jankowski_2020_deep}, achieving efficient encoding and facilitating effective transmission across subsequent channels.
    
        These principles from communication models provide a framework for constructing and optimizing neural networks. Information theory, developed by Shannon \cite{jones_2000_information}, introduces the concept of entropy to analyze information processes through the lens of quantification. Entropy, a measure of uncertainty in signal processing, highlights the intrinsic limitations of traditional communication systems and guides the design of more efficient architectures.
    
        The integration of information theory into neural networks has been a focus of research, with MacKay et al. \cite{mackay_2003_information} proposing a channel model based on neural mechanisms. Furthermore, Sharma et al. \cite{sharma_2021_dagsurv} introduced fiducial coding within variational autoencoders, employing Shannon's first theorem to balance the entropy of encoding against its length, ensuring a distortion-free transmission.
    
        Building on these concepts, Tishby and Zaslavsky \cite{tishby_2000_the} explored the architecture of multilayer perceptrons to understand how networks manage and utilize information. Their findings suggested that networks prioritize capturing pertinent information early in the processing chain, which is then synthesized in subsequent layers. Inspired by these insights and the methodologies of DAGSurv \cite{sharma_2021_dagsurv}, our approach models the feature extraction phase of a single modality as a source coding problem. This strategy allows for the efficient compression of information and enhances overall model performance by training the feature extraction in concert with fusion and detection networks. By selectively capturing and utilizing effective features while minimizing redundancy, our model improves both efficiency and accuracy.
    
        Importantly, we incorporate entropy-based metrics from information theory to quantify the data processed by each network layer, aiming to restrict entropy fluctuations and thus streamline the optimization pathway. This approach not only enhances the efficiency of the network but also aligns with the interpretability goals essential for applications in complex environments such as intelligent driving, where understanding model decisions is crucial for safety and reliability.
    
        Our ongoing research continues to refine these techniques, seeking to fully leverage the principles of information theory to improve the interpretability and functionality of deep learning models in high-stakes applications.

    \subsection{Optimizer of Neural Network Training Strategy}
    
        Optimizing neural network training involves fine-tuning parameters to minimize the loss function, which is essential for enhancing model performance. This process encompasses evaluating performance metrics across the entire training dataset and integrating additional regularization to prevent overfitting. The predominant strategies for optimization are categorized into three groups: gradient descent methods, momentum optimization methods, and adaptive learning methods.
    
        \textbf{Gradient Descent Methods:} The gradient descent strategy involves updating the network parameters iteratively in the opposite direction of the gradient of the loss function concerning the parameters. Among the variants, Batch Gradient Descent (BGD)\cite{hinton2012neural} and Stochastic Gradient Descent (SGD)\cite{bottou2012stochastic} represent two ends of the spectrum. BGD processes the entire dataset in one go, ensuring stable convergence but at a high computational cost. In contrast, SGD updates parameters more frequently using individual training samples, which introduces noise into the training process but allows for faster convergence. The mini-batch gradient method strikes a balance by processing subsets of the training data, effectively combining the stability of BGD with the efficiency of SGD \cite{ruder2016overview}.
    
        \textbf{Momentum Optimization Methods:} These methods incorporate concepts from physics, such as velocity and momentum, to accelerate gradient descent. Techniques like Momentum and Nesterov Accelerated Gradient (NAG) help to stabilize the updates. Momentum optimization calculates an exponentially weighted average of past gradients and continues to move in that direction, which amplifies the speed of descent in consistent gradient directions while damping oscillations. NAG, an enhancement over Momentum, adjusts the gradient calculation by considering the position where parameters are likely to be in the next step, thus providing foresight into future updates \cite{dozat2016incorporating}.
    
        \textbf{Adaptive Learning Rate Methods:} Determining the optimal learning rate is a critical yet challenging aspect of training neural networks. Adaptive learning rate methods adjust the learning rate dynamically based on the training progress. Popular algorithms like AdaGrad, RMSProp, Adam, and AdaDelta each provide unique mechanisms for adjusting the learning rate. AdaGrad adjusts the learning rate inversely proportional to the square root of the sum of all previous squared gradients, making it suitable for sparse data. RMSProp modifies AdaGrad by incorporating a moving average of squared gradients to provide more responsive adjustments. Adam combines the benefits of Momentum and RMSProp, adjusting learning rates based on both the first and second moments of gradients, thus ensuring efficient convergence across various conditions \cite{le2011optimization,zaheer2019study}. Adam also integrates bias corrections to address initialization bias, making the optimizer robust to different initialization schemes.
    
        These optimization strategies are essential tools for training deep neural networks efficiently and effectively. By carefully selecting and tuning these methods, practitioners can significantly enhance the performance and convergence speed of neural network models, which is critical for applications requiring real-time processing and high-accuracy outcomes such as in autonomous driving or medical diagnostics.

\section{Theory}

    Efficient fusion of multimodal data necessitates the compression of non-essential information to ensure that only relevant features contribute to subsequent tasks within a network. This concept, akin to distortion-limited encoding in communication models, involves the selective removal of less frequently occurring source symbols to enhance the efficiency of information transmission.

    In practical terms, source symbols that are infrequent within the dataset might not significantly contribute to the outcome of analytical tasks. Their removal during the compression phase can substantially increase data recovery rates at the decoding end, thus optimizing transmission efficiency without compromising the integrity of the data. This principle is directly applicable in neural networks designed for processing complex information, where removing redundant or non-informative data can lead to improved computational efficiency and faster processing times \cite{jones_2000_information}.

    Entropy, a fundamental concept in information theory, serves as a quantitative measure of information randomness or uncertainty. It is inversely related to the predictability and usefulness of the information. By implementing an entropy calculation method, we can quantitatively assess the quality of information retained or discarded during the compression process. This method not only aids in maintaining the fidelity of information but also ensures that the encoding process is efficient and free from unnecessary redundancies.

    To ensure consistent quality and prevent abrupt changes in data quality, it is crucial to regulate the rate of entropy change across different layers of the neural network. Maintaining a steady entropy gradient prevents sudden distortions in data representation, thereby preserving the integrity of information throughout the learning and decision-making processes. This approach underlines the importance of strategic data handling and emphasizes the balance between data compression and preservation in achieving optimal performance in neural network applications.

    \subsection{Entropy Expectation of Neural Network Layers}

        \begin{figure}[H]
          \centering
          \includegraphics[width=0.8\linewidth]{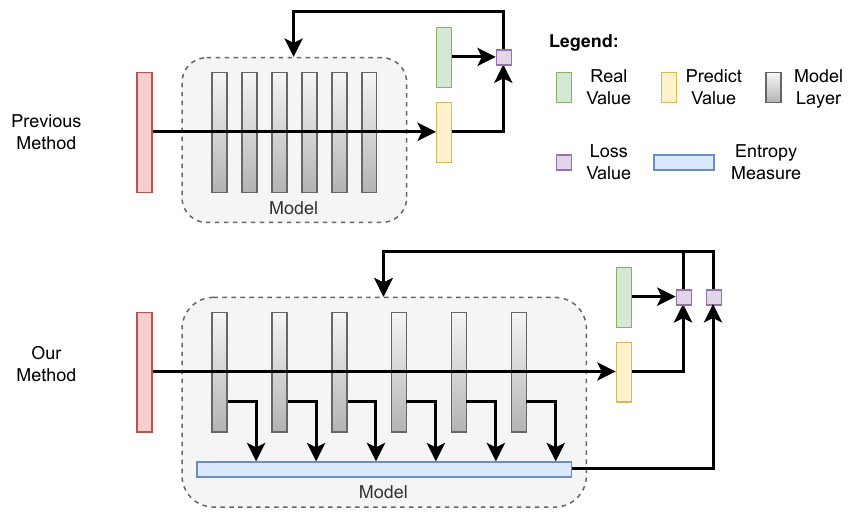}
          \caption{Comparison between previous training method and our training method.}
          \label{method-structure}
        \end{figure}
        
        The training of neural networks involves an iterative process of mapping inputs to their expected outputs, a concept essential for machine learning \cite{liu2017survey}. Initially, this mapping relationship is typically weak; however, as training progresses with data, this relationship is reinforced, enhancing the network's feature extraction—essentially, its ability to compress information. Despite these advancements, the opacity of such "black-box" models often obscures the internal workings of the information compression process, making it challenging to guide the optimization of network parameters effectively \cite{buhrmester2021analysis}.
        
        To demystify the mechanics behind information compression in neural networks, we draw parallels with source coding in communication systems. Here, feature compression is akin to distortion-limited encoding, designed to ensure comprehensive data extraction that is pertinent to subsequent tasks. This approach ensures that the encoding process retains only the most critical information, thereby enhancing processing efficiency.
        
        By incorporating the concept of information entropy, we measure the information content processed by the network. In systems where bandwidth remains constant, reducing entropy across the data transmission process signifies enhanced efficiency of information handling \cite{zou2022novel}. This reduction in entropy is indicative of an increase in the degree of encoding within distortion-limited encoders, a principle that is similarly applicable to feature compression networks.
    
        Notably, feature compression networks frequently exhibit a repetitive structural design, such as the repeated linear layers found in the SECOND network \cite{yan2018second}. This design suggests that each layer likely possesses a consistent capability for compressing data, thereby maintaining steady bandwidth throughout the network's operation. The primary expectation from such a setup is a gradual, consistent decrease in the information entropy output by each layer, reflecting a stable and efficient compression process.
    
        With these insights, it is possible to tailor the network's parameters to foster a consistent entropy reduction across layers by deploying an entropy loss function. This strategy allows deeper penetration into the "black-box" nature of the neural networks, enabling more precise and effective training optimization. Through this methodology, we not only enhance the interpretability of neural networks but also improve their efficiency and reliability in tasks demanding high levels of data compression and transformation.

    \subsection{Probabilistic Modeling for Information}

        The effective computation of loss functions in neural networks, particularly those tailored for information compression, hinges on the ability to accurately estimate entropy at each layer. This estimation is intricately tied to the probabilistic modeling of the outputs from these layers, necessitating a rigorous approach to understanding and analyzing the data distributions involved.
    
        A common strategy in information compression utilizes convolutional networks, which are highly effective due to their structured approach to handling spatial hierarchies in data \cite{zou2022novel}. In the context of probabilistic modeling, we treat the outputs from convolutional layers as samples from a multidimensional continuous random variable. Specifically, the feature channels $\tilde{X} = \{x_1, x_2, \ldots, x_i\}$ produced by the convolutional processes are considered as instances of the random variable $X$, where $i$ denotes the number of channels and $d$ represents the dimensionality of each channel.
    
        This probabilistic framework allows for the conceptualization of network outputs at each layer as continuous random variables $X$, with the specific outputs $x_i$ serving as the sampled data points. This perspective is not limited to convolutional neural networks; it can be readily adapted to other types of neural architectures. By applying this model across various network designs, we can extend the utility of probabilistic modeling to encompass a wide range of neural network structures.
    
        The power of this approach lies in its ability to provide a unified method for quantifying the entropy of data distributions within neural networks. By estimating the entropy across different layers and network configurations, we can gain deeper insights into the information dynamics within the network. This, in turn, aids in optimizing the neural network by focusing on reducing entropy in a controlled manner, thereby enhancing the overall efficiency and effectiveness of the network in tasks involving complex data compression and feature extraction.
    
        In summary, adopting a probabilistic modeling framework for analyzing and interpreting neural network outputs provides a robust basis for entropy estimation. This methodology not only enhances our understanding of the internal mechanisms of neural networks but also opens up avenues for refining network training strategies and improving model performance across various applications.

    \subsection{Entropy Calculation}
        The challenge of estimating the entropy at each layer of a neural network boils down to evaluating the entropy based on the probability distribution of the unknown continuous random variable $X$.
    
        To address this, we leverage the concept of differential entropy, which extends Shannon's classical definition of entropy to continuous probability distributions \cite{jones_2000_information}. The differential entropy $h(X)$ for a random variable $X$, with a probability density function $f$ over its domain, is defined as:
        
        \begin{equation}
            h(X) = -\int f(x) \log f(x) \, dx
        \end{equation}
        
        However, without prior knowledge of the exact probability distribution, and with only a limited set of sample values available from this distribution, estimating $f(x)$ directly is impractical. In such cases, the K-Nearest Neighbor (KNN) Entropy Estimation Method provides a viable alternative \cite{van1988generalized}. This method approximates the differential entropy by considering the spatial distribution of sample points within the data space.
        
        In the KNN approach, each sample point is considered within a d-dimensional hypersphere, with the radius defined by the distance to the nearest neighbor. Assuming uniform distribution within this volume, the probability of each point can be approximated by $1/n$, where $n$ is the total number of samples. Yet, given the likely non-uniform distribution of data in practice, this method adjusts the estimated probability density based on local sample density, as follows:
        
        \begin{equation}
            p(x_i) = \left[(n-1) \cdot r_d(x_i)^d \cdot V_d\right]^{-1}
        \end{equation}
        
        Here, $r_d(x_i)$ is the Euclidean distance from the sample point $x_i$ to its nearest neighbor, and $V_d$ represents the volume of the unit sphere in d dimensions. The entropy estimate for the distribution is then calculated by summing the logarithmic probabilities adjusted by the Euler-Mascheroni constant $\gamma$, approximately 0.5772:
    
        \begin{equation}
            H(X) = \frac{1}{n} \sum_{i=1}^n \left[-\log p(x_i)\right] + \gamma
        \end{equation}
    
        For more robust estimation, this method can be extended to consider distances to the k-th nearest neighbor, enhancing the accuracy especially in sparse data regions:
    
        \begin{equation}
            H(X,k) = -\psi(k) + \psi(n) + \log V_d + \frac{d}{n} \sum_{i=1}^n \log r_{d,k}(x_i)
        \end{equation}
    
        Here, $\psi$ is the digamma function, and $r_{d,k}(x_i)$ indicates the distance to the k-th nearest neighbor. Notably, this refined entropy calculation, $H(X,k)$, closely approximates $H(X)$ when $k=1$. We utilize this entropy measure, $H(X)$, to derive the entropy change $\Delta H$ across network layers, with $\Delta H_n = H_{n+1} - H_n$, where $n$ is the layer index. This precise measurement of entropy changes allows for enhanced optimization of the network's information processing capabilities, ensuring efficient learning and data representation.
    
    \subsection{Loss Functions for Information Compression Network}

        In the realm of information compression networks, the overarching goal is to transmit data efficiently while minimizing redundancy and irrelevant information. To this end, we have developed two specific loss functions based on the expected behavior of the information transmission process within such networks. These functions, $L_1$ and $L_2$, are designed to optimize the network by focusing on the steadiness and directionality of information change, respectively.
    
        \textbf{Entropy Variance Loss Function ($L_1$):} This function is aimed at ensuring a stable change in entropy across the network layers. It measures the variance of the entropy changes $\Delta H$ against their mean $\widehat{\Delta H}$, with an ideal variance of zero, indicating perfect stability:
    
        \begin{equation}
            L_1 = \frac{\sum_{n=1}^N (\Delta H_n - \widehat{\Delta H})^2}{N}
        \end{equation}
    
        Here, $N$ represents the total number of layers, and $n$ is the index of each layer. The mean entropy change $\widehat{\Delta H}$ is calculated as the average of entropy changes across all layers, thus providing a benchmark for assessing deviations in individual layers' performance.
    
        \textbf{Entropy Direction Loss Function ($L_2$):} In contrast to $L_1$, the loss function $L_2$ concentrates on the directionality of the entropy change, specifically aiming to reduce entropy consistently across the network:
    
        \begin{equation}
            L_2 = -\sum_{n=1}^N \Delta H_n^2
        \end{equation}
    
        This formulation inherently promotes a decrease in entropy values, thereby encouraging information compression and enhancing the efficiency of the network.
    
        Collectively referred to as \textit{Entropy Loss}, the combination of $L_1$ and $L_2$ offers a comprehensive framework for tuning the network's behavior towards optimal information processing. This dual approach not only enhances the interpretability of the network by clarifying the dynamics of information change but also addresses specific challenges associated with the compression and transmission of data.
    
        However, it is important to note that the influence of Entropy Loss is predominantly effective in network layers that exhibit repetitive structures and are dedicated to specific tasks closely aligned with principles from communication theory. This specificity can be seen as both a strength, in terms of targeted optimization, and a limitation, as it may not generalize across all types of network architectures or tasks.
    
        As such, while Entropy Loss significantly contributes to the network's ability to process and compress information efficiently, it should be viewed as a complement to other loss functions ($L$) that directly target the end goals of the network. This layered approach to loss function design ensures that while we optimize for information efficiency, we also remain aligned with the ultimate performance objectives of the network.

\section{Experiments}

    \textbf{Dataset.} Our research utilizes the KITTI dataset, a cornerstone in the field of computer vision, particularly for intelligent driving applications \cite{geiger_2012_are, geiger2013vision}. Developed collaboratively by the Karlsruhe Institute of Technology in Germany and the Toyota Institute of Technology in America, this dataset offers a comprehensive range of real-world scenarios captured from urban areas, villages, and highways. It includes a rich assembly of multimodal data such as lidar point clouds, GPS data, right-hand color camera data, and grayscale camera images. The KITTI dataset is meticulously organized into training and test sets, containing 7,481 and 7,518 samples respectively, providing a robust basis for evaluating the performance of computer vision algorithms in realistic settings.

    \textbf{Implementation Details.} The experimental framework for our study was established using the Nvidia RTX 3090 graphics processing unit, renowned for its computational efficiency and suitability for high-demand machine learning tasks. We constructed our models within the PyTorch ecosystem, leveraging its extensive suite of deep learning tools and its inherent flexibility in handling dynamic computational graphs \cite{mmdet3d2020}. PyTorch's auto-differentiation capability significantly simplifies the implementation of complex models by automating the calculation of gradients, an essential feature that enhances the development of sophisticated neural network architectures. Furthermore, our models are developed on MMDetection3D, an open-source toolbox explicitly designed for 3D object detection, which extends PyTorch's capabilities into three-dimensional data processing. This framework supports a broad range of 3D detection algorithms, making it an invaluable resource for advancing research in intelligent driving systems.

    \subsection{Evaluate the Influence of Entropy Loss}
        \begin{figure}[ht]
          \centering
          \subfloat[Distribution without Entropy Loss.]{
            \includegraphics[width=\linewidth]{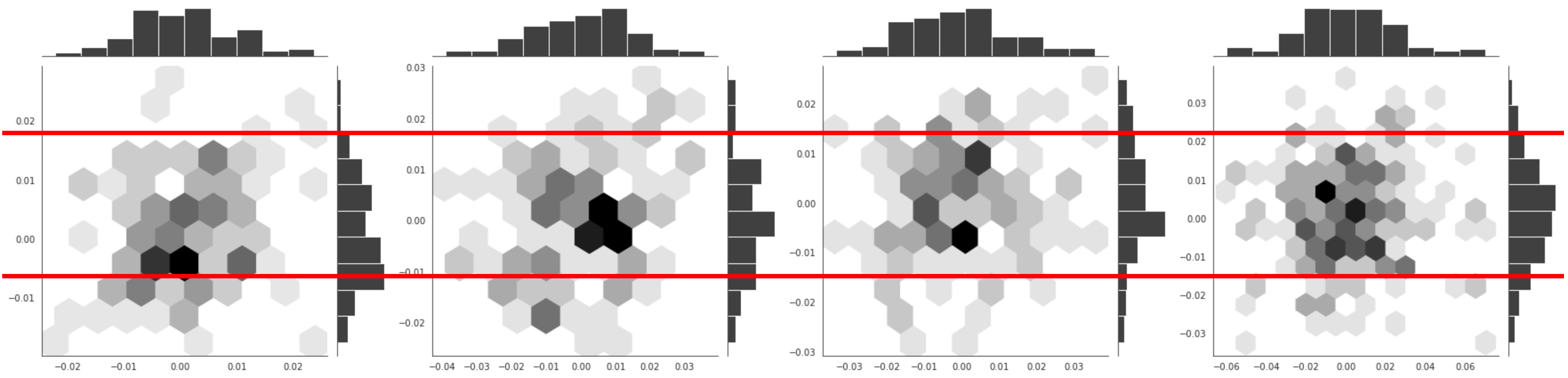}
          }
          \newline
          \subfloat[Distribution with Entropy Loss.]{
            \includegraphics[width=\linewidth]{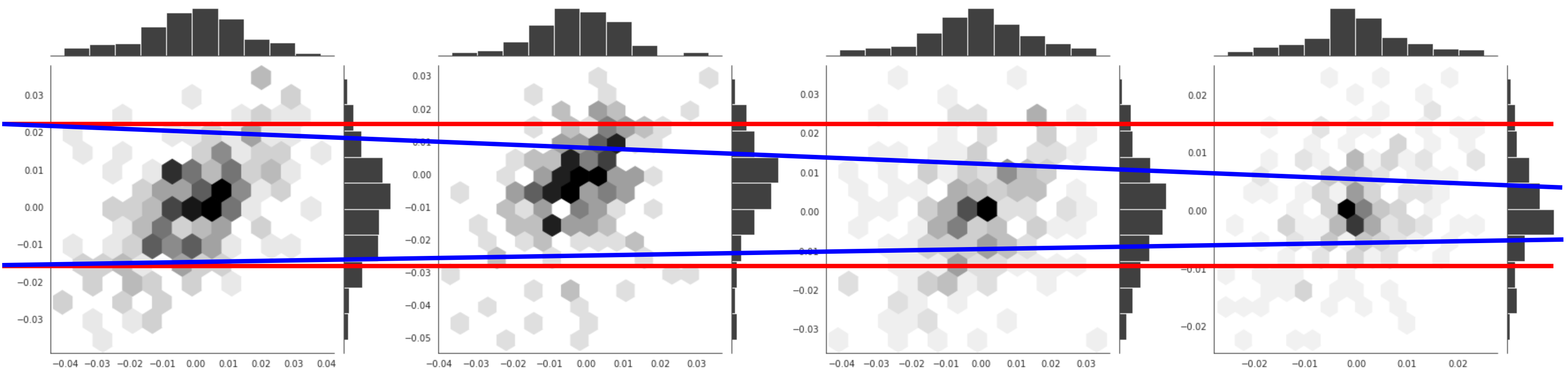}
          }  
          \caption{Comparison between the distribution of feature vectors in feature space generated by SECOND with or without Entropy Loss.}
          \label{distribution-figure}
        \end{figure}
        
        The impact of Entropy Loss on the feature distribution within neural networks is profound and merits detailed exploration. To illustrate the effect more intuitively, we analyzed the output of each layer of a model trained with and without the Entropy Loss function. Utilizing Principal Component Analysis (PCA) \cite{abdi2010principal} to reduce dimensionality, we produced the visualizations shown in Figure~\ref{distribution-figure}. These images highlight the feature compression process, represented by the red and blue lines.
    
        In scenarios without Entropy Loss, discerning a consistent pattern or rule in the distribution of output feature samples from each network layer proves challenging. This variability can lead to inefficiencies in learning and model generalization. Conversely, the integration of Entropy Loss stabilizes the transformation of data throughout the network layers, as evidenced by the more orderly and predictable distributions depicted in the figure. This stabilization suggests that Entropy Loss not only enhances the compressibility of features but also aligns them more closely with the underlying structure dictated by the target functions of the network.
    
        These findings underscore the dual benefits of incorporating Entropy Loss into neural network training protocols. Firstly, it promotes a more structured and interpretable feature space, which is crucial for the robustness and reliability of learning outcomes. Secondly, the orderly progression of feature transformation across layers implies a more efficient encoding of information, which can be particularly advantageous in tasks requiring precise and rapid processing of complex data streams.

    \subsection{Comparison with Entropy Loss on Training Process}
        \begin{figure}[ht]
          \centering
          \includegraphics[width=\columnwidth]{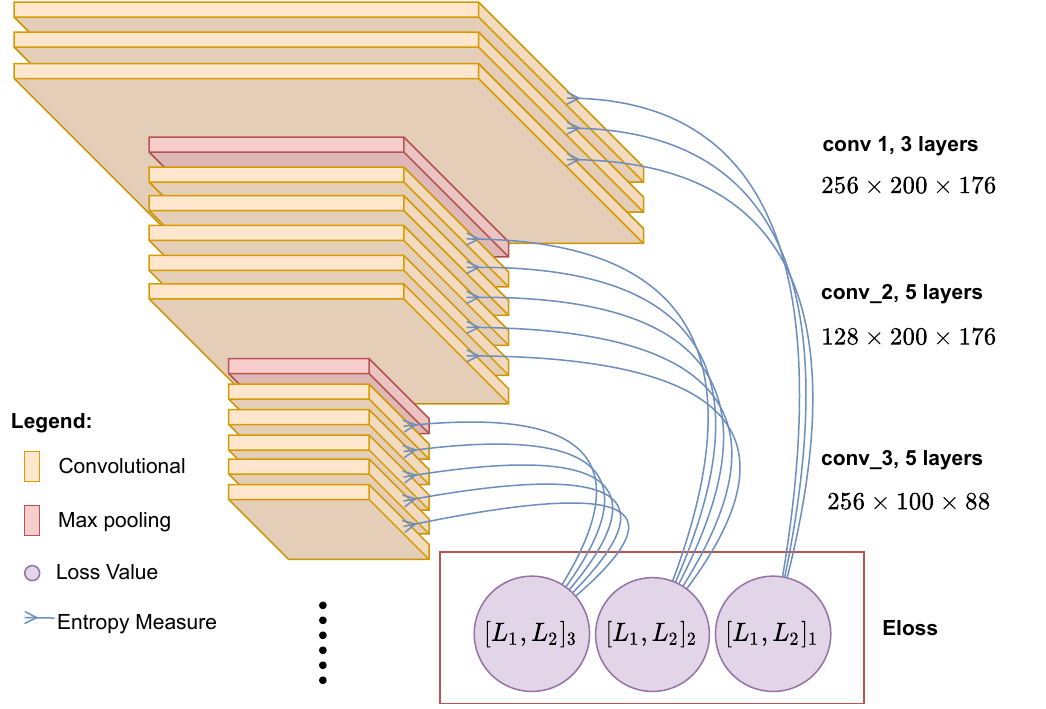}
          \caption{A demonstration where Entropy Loss is applied in SECOND backbone \cite{yan2018second}}
          \label{real-impl-figure}
        \end{figure}

        \begin{figure*}[h]
          \centering
          \subfloat[Car]{
            \includegraphics[width=0.32\textwidth]{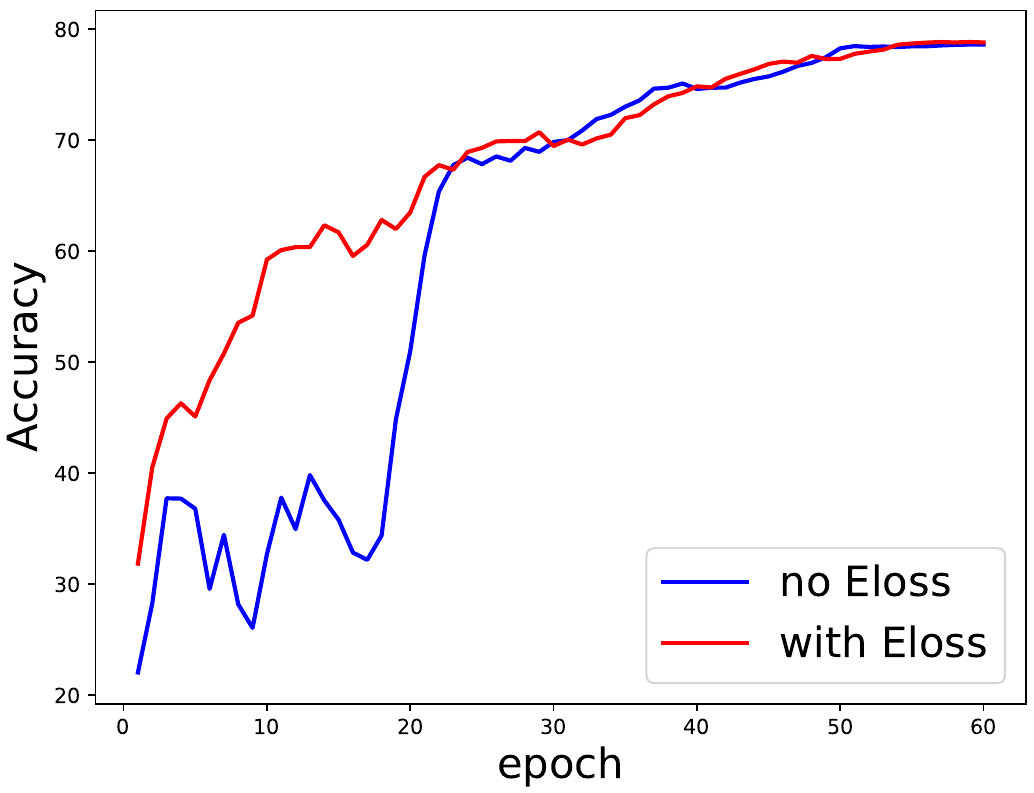}
          }
          \hfill
          \subfloat[Cyclist]{
            \includegraphics[width=0.32\textwidth]{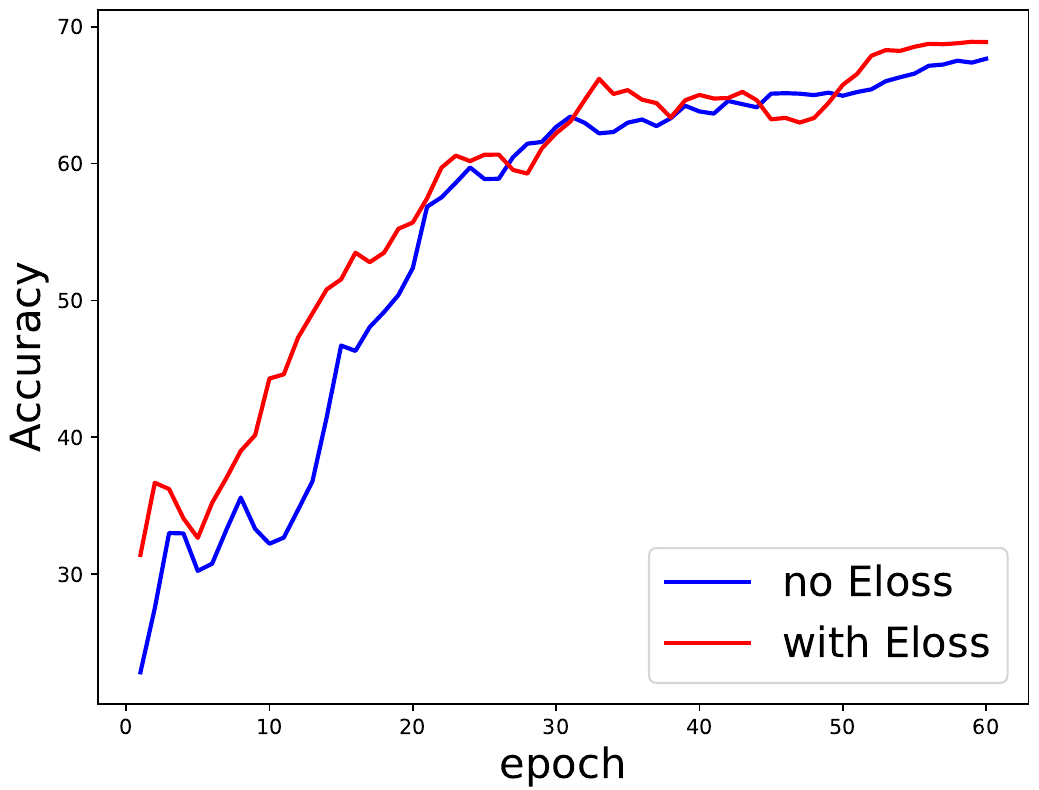}
          } 
          \hfill
          \subfloat[Pedestrian]{
            \includegraphics[width=0.32\textwidth]{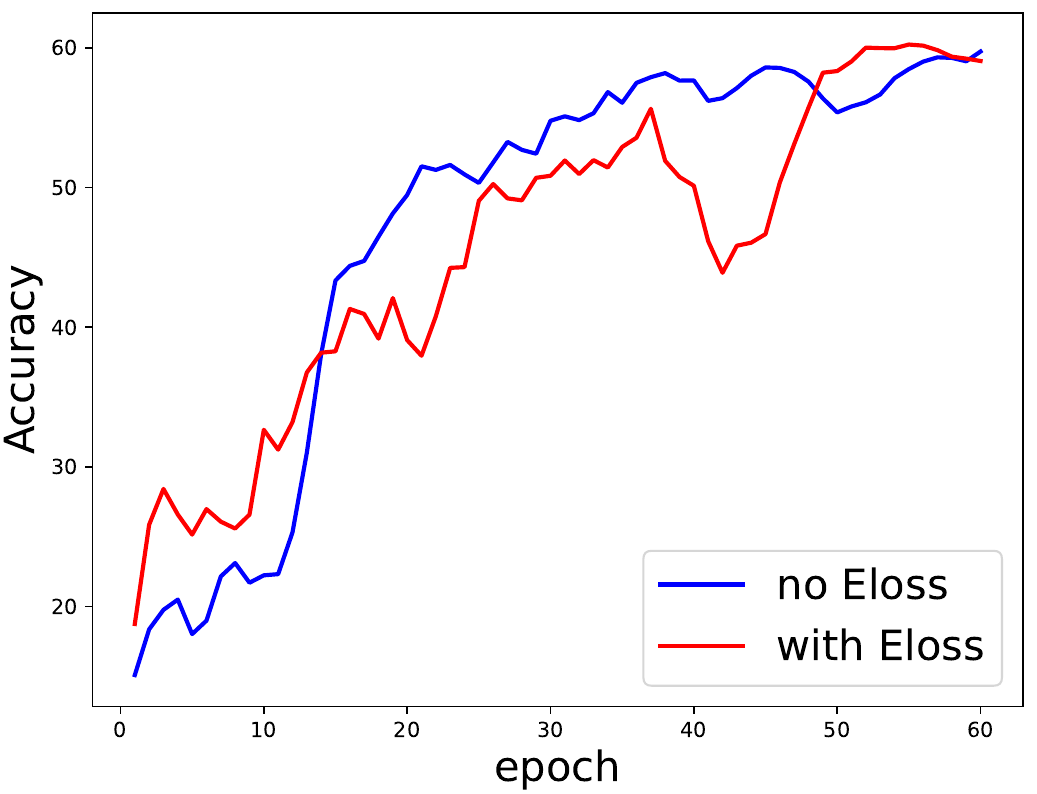}
          } 
          \caption{Convergences Curves of the model accuracy on validation set for SECOND \cite{yan2018second} and SECOND with Entropy Loss on 3 Class: Car, Cyclist and Pedestrian.}
          \label{comparison-figure}
        \end{figure*}
        
        \begin{minipage}{0.48\textwidth}
            \makeatletter\def\@captype{table}
                \label{evaluation-table-mean}
                \centering
                \begin{tabular}{cccc}
                    \toprule
                    Entropy Loss     &  Car  & Cyclist  & Pedestrian  \\
                    \midrule
                    \midrule
                    & 58.49  & 53.02 & 45.52 \\
                    \Checkmark & 65.49  & 56.17 & 43.97 \\
                    \midrule
                    Delta & +\textbf{7.11}  & +\textbf{3.15} & -1.55 \\
                    \bottomrule
                \end{tabular}
            \caption{The avearge accuracy of the model on validation set during training process.}
        \end{minipage}
        \hspace{8px}
        \begin{minipage}{0.48\textwidth}
            \makeatletter\def\@captype{table}
            \label{evaluation-table-r2}
            \centering
            \begin{tabular}{cccc}
                \toprule
                Entropy Loss     &  Car  & Cyclist  & Pedestrian  \\
                \midrule
                \midrule
                & 0.738  & 0.848 & 0.826 \\
                \Checkmark & 0.919 & 0.863 & 0.745 \\
                \midrule
                Delta & +\textbf{0.181} & +\textbf{0.015} & -0.081 \\
                \bottomrule
            \end{tabular}
            \caption{The R-squared ratio of the log regression of the model accuracy on validation set.}
        \end{minipage}
    
        To rigorously evaluate the impact of Entropy Loss on the training dynamics, we employed a unimodal 3D object detection model, SECOND \cite{yan2018second}, trained over 60 epochs on the KITTI dataset. The structure of the modified SECOND model incorporating Entropy Loss is depicted in Figure~\ref{real-impl-figure}. Results, visualized in Figure~\ref{comparison-figure}, highlight the smoothed accuracy curves calculated by averaging the five closest values for each data point.

        Table~\ref{evaluation-table-mean} presents the average accuracy throughout the training process, indicating an overall increase of $2.9\%$ in mean accuracy, which will be discussed further in the context of test set performances. Additionally, Table~\ref{evaluation-table-r2} details the $R^2$ ratio from the logarithmic regression analysis of each accuracy curve, where a notable improvement of $0.169$ average increment in precision stability is observed with Entropy Loss. This suggests that Entropy Loss not only improves overall model accuracy but also contributes to more consistent learning outcomes across different iterations.
    
        The analysis across different classes within the KITTI dataset—Car, Cyclist, and Pedestrian—reveals that Entropy Loss generally enhances detection precision, particularly in categories with ample training data. This enhancement underscores the effectiveness of Entropy Loss in refining the model's ability to generalize from training data, a crucial attribute for robust real-world applications.

    \subsection{Comparison between Different Models}
    
        \begin{table*}[ht]
          \centering
          \resizebox{\textwidth}{!}{
              \begin{tabular}{ccccccccccc}
                \toprule
                & & & Car & & & Cyclist & & & Pedestrian & \\
                \cmidrule(r){3-5} \cmidrule(r){6-8} \cmidrule(r){9-11}
                Model     &  Entropy Loss  &  Easy  & Moderate  & Hard &  Easy  & Moderate  & Hard &  Easy  & Moderate  & Hard \\
                \midrule
                \midrule
                SECOND\cite{yan2018second} &  & 82.35 & 73.35  & 68.59 & 70.89 & 56.72 & 50.68 & 50.75 & 40.76 & 36.96\\
                 & \Checkmark & 82.68 & 73.67  & 67.21 & 71.99 & 58.00 & 50.94 & 50.49 & 41.16 & 37.43 \\
                \cmidrule(r){2-11}
                 & Delta & +\textbf{0.33} & +\textbf{0.32} & -1.38 & +\textbf{1.1} & +\textbf{1.28} & +\textbf{0.26} & -0.26 & +\textbf{0.4} & +\textbf{0.47} \\
                 \midrule
                +ResNet\cite{he2016deep} &  & 80.29 & 67.37 & 60.94 & 75.70 & 52.37 & 46.10 & 39.64 & 31.13 & 28.95 \\
                 & \Checkmark * & 77.62 & 64.92 & 60.36 & 71.47 & 55.79 & 49.64 & 44.85 & 35.60 & 32.66 \\
                 \cmidrule(r){2-11}
                 & Delta & -2.67 & -2.45 & -0.58 & -4.23 & +\textbf{3.42} & +\textbf{3.54} & +\textbf{5.21} & +\textbf{4.47} & +\textbf{3.71}\\
                 \midrule
                +Correlation\cite{zheng2022multi} &  & 73.47 &  62.47  & 57.99 & 63.08 & 49.55 & 44.33 & 42.46 & 35.11 & 32.16 \\
                +GNN\cite{scarselli2008graph} & \Checkmark * & 67.33 & 58.70  & 54.13 & 57.16 & 46.36 & 41.54 & 45.02 & 36.39 & 33.29  \\
                 \cmidrule(r){2-11}
                +FPN\cite{lin2017feature} & Delta & -6.14 & -3.77 & -3.86 & -5.92 & -3.19 & -2.79 & +\textbf{2.56} & +\textbf{1.28} & +\textbf{1.13} \\
                \bottomrule
                * Adding Entropy Loss to SECOND only.
              \end{tabular}
            }
            \caption{Compare between the accuracy on test set for 3 different models with or without Entropy Loss on 3 Class: Car, Cyclist, and Pedestrian, after 40 epochs train.}
          \label{comparision-table}
        \end{table*}
        
        To further evaluate the influence of Entropy Loss on model precision, we applied models equipped with this modification to the KITTI test set. The compiled results, detailed in Table~\ref{comparision-table}, illustrate varying degrees of performance enhancement across different model configurations and object classes.

        The analysis demonstrates a modest overall improvement in the SECOND model's accuracy by approximately $0.28\%$ after 40 epochs of training with Entropy Loss. This increment, although slight, underscores the potential for Entropy Loss to refine the model's predictive accuracy.
    
        Additional complexity arises when integrating Entropy Loss with advanced architectures like SECOND+ResNet, where information from dual modalities—point cloud and image—is processed. While the detection accuracy for Cyclist and Pedestrian classes saw improvements exceeding $3\%$, a notable decrease occurred in Car detection accuracy. This trend becomes more pronounced in more complex configurations such as SECOND+ResNet+Correlation\cite{yan2018second,he2016deep,zheng2022multi}+GNN\cite{scarselli2008graph}+FPN\cite{lin2017feature}, where only the Pedestrian detection accuracy improved. 
    
        These findings suggest that the benefits of Entropy Loss are most pronounced in simpler network layers and become diluted as the model's complexity increases. This phenomenon may indicate that Entropy Loss's effectiveness is contingent upon the model's ability to leverage structured, entropy-influenced learning, which may be overshadowed in highly complex models handling diverse data types.

\section{Conclusion}

In this work, we introduced Entropy Loss, a novel concept designed to amplify the interpretability of feature compression networks, drawing upon principles from communication systems. This loss function is meticulously crafted by aligning network-layer outputs with predefined expectations of information change, which are rooted in the theories of source coding. Our implementation of Entropy Loss aims to steer network optimization towards these expectations, effectively guiding the training process.

Our empirical investigations across three distinct aspects reveal that incorporating Entropy Loss into the training of 3D object detection models not only enhances training efficiency but also significantly improves model interpretability. These improvements are crucial for applications within intelligent driving systems, where understanding the model's decision-making process is as important as its predictive accuracy.

However, our findings also underscore certain limitations of Entropy Loss, particularly its impact on models where only a small portion of the network architecture is amenable to influence by this loss function. In such cases, Entropy Loss may inadvertently impede the training process, leading to suboptimal performance. This observation highlights the nuanced role of Entropy Loss in complex neural network architectures and suggests a potential area for further investigation.

Moving forward, our research will delve deeper into these limitations, seeking to refine the application of Entropy Loss and extend its benefits more uniformly across various network configurations. We aim to continue exploring the interpretability enhancements that Entropy Loss can provide, particularly in the context of intelligent driving, where the stakes and complexities are notably high. Through these efforts, we anticipate developing more robust and transparent models capable of driving advancements in autonomous vehicle technologies.


\bibliographystyle{plain} 
\bibliography{reference} 

\end{document}